\documentclass{Interspeech2024}
\usepackage{multirow}

\interspeechcameraready

\title{A Comparative Analysis of Bilingual and Trilingual Wav2Vec Models for Automatic Speech Recognition in Multilingual Oral History Archives}

\name{Jan}{Lehečka}
\name{Josef V.}{Psutka}
\name{Luboš}{Šmídl}
\name{Pavel}{Ircing}
\name{Josef}{Psutka}
\address{Department of Cybernetics, University of West Bohemia Pilsen, Czech Republic}
\email{\{jlehecka,psutka\_j,smidl,ircing,psutka\}@kky.zcu.cz}

\keywords{speech recognition, bilingual models, trilingual models, oral history archives}

\begin{document}

\maketitle

\begin{abstract}
    
    In this paper, we are comparing monolingual Wav2Vec 2.0 models with various multilingual models to see whether we could improve speech recognition performance on a unique oral history archive containing a lot of mixed-language sentences. 
    Our main goal is to push forward research on this unique dataset, which is an extremely valuable part of our cultural heritage.
    Our results suggest that monolingual speech recognition models are, in most cases, superior to multilingual models, even when processing the oral history archive full of mixed-language sentences from non-native speakers.
    We also performed the same experiments on the public CommonVoice dataset to verify our results.
    We are contributing to the research community by releasing our pre-trained models to the public.

\end{abstract}

\section{Introduction}

One of the most valuable possessions mankind has is the heritage from previous generations in the form of historical documents and recordings. This heritage preserves the memory of humanity which should not be forgotten.
Since collections of historical documents and recordings can grow huge in size, one of the key challenges of our age is to preserve and curate this heritage and make it as accessible and searchable to the public and researchers as possible in order to enable advanced analyzing, studying, and learning of new lessons from our history.

In this paper, we focus mainly on one very important and unique oral history archive: MALACH. It is an audiovisual archive initially collected in the 1990s to preserve the authentic memories of Holocaust survivors. This archive stores vast and extremely valuable testimonies from our recent history recorded through audiovisual interviews.
Today, these interviews are stored in the Visual History Archive (VHA) at the Shoah Foundation Institute (SFI) at the University of Southern California (USC)\footnote{\url{https://vha.usc.edu}}, along with other interviews with witnesses to the history of the entire 20th century (more than 56k interviews). The Holocaust part of the archive contains testimonies in 32 languages of the personal memories of people who survived the World War II Holocaust. In this paper, we denote this archive and derived datasets as \textit{MALACH} (Multilingual Access to Large Spoken Archives) after the name of the first project that began in 2001 and laid the foundations for work on this unique archive.\footnote{\url{https://malach.umiacs.umd.edu}}

The natural way how to process large oral history archives to make the content more accessible is to transcribe the speech using an Automatic Speech Recognition (ASR) system.
Although ASR systems have improved rapidly in the last years \cite{baevski2020wav2vec,babu22_interspeech,radford2022whisper}, automatically transcribing interviews from the MALACH is still a challenge since the interviews contain spontaneous speech full of disfluencies, emotional excitements, mixed-language sentences, and heavy accents, and are often influenced by the high age of speakers (the average age of all speakers at the time of recording was about 75 years) \cite{10.1007/978-3-030-87802-3_50,lehecka23_interspeech}. 

The large number of mixed-language sentences, mainly the German phrases present in interviews across all other languages\footnote{E.g. a sentence from an English interview: \textit{“I said yes, Herr Lagerführer, ich habe gehalten Hemdentashen, so he give me two more.”}}, together with the natural multilingualism of this archive (32 languages with many non-native speakers) motivated us to study multilingual aspects of this archive and try to answer some interesting questions:
Would for example adding some German speech data into the training process of the English ASR model solve the problem of transcribing mixed-language sentences?
More generally, would the bilingual pre-trained models be more suitable for this task than monolingual models? And how about trilingual models  -- could they be even more suitable? Ultimately, would a large-scale massively multilingual ASR model transcribe the interviews better than a set of per-language monolingual ASR models?
How much will results from these approaches differ?

To answer these questions, in this work, we present a comparative analysis over a large set of experiments with different language combinations in both pre-training and fine-tuning of ASR systems based on Wav2Vec 2.0 models \cite{baevski2020wav2vec} while simplifying the problem only to 3 languages: English, German, and Czech.
Since some of the MALACH datasets are not publically available, we fine-tuned, evaluated, and compared our models also on another well-known and publicly available dataset -- CommonVoice \cite{commonvoice:2020} -- to see if our findings are applicable also to other multilingual datasets than oral history archives.

\section{Related work}
\label{sec:relwork}
The original MALACH project took place between 2001 and 2006.
The WER of the ASR systems developed within the project reached 39.40\% for English~\cite{Byrne_2004} and 38.57\% for Czech~\cite{Psutka_2005} by the end of the project in 2006. German part was not processed with ASR at that time. Even after the project finished, the efficiency of the ASR systems has continuously improved using new approaches, so in 2011 the WER of 27.11\% was achieved for Czech recordings~\cite{Psutka_2011}. New training methods based on DNN brought further improvement of WER (21.70\% for English~\cite{picheny_2019} and 19.11\% for Czech~\cite{Svec_2017}). The best WER without using end-to-end approaches reached 17.85\% for English and 14.65\% for Czech \cite{10.1007/978-3-030-87802-3_50}.
After the introduction of end-to-end Transformer-based audio models, \cite{10.1007/978-3-031-16270-1_25} reported a significant improvement for the Czech dataset (WER=10.48\%) and \cite{picheny23_interspeech} for English (WER=13.5\%). The last state-of-the-art results we are aware of are 12.88\% for English, 17.08\% for German, and 8.43\% for Czech reported in \cite{lehecka23_interspeech}.

All recent works on MALACH datasets (and a majority of other ASR datasets) use either monolingual models or large massively multilingual models (pre-trained on 100+ languages). However, no attention has been paid to bilingual and trilingual ASR models in the literature so far.

\subsection{Wav2Vec 2.0 model}
After the introduction of the Transformer architecture \cite{vaswani2017attention}, a new era of AI began. Not long after that, Transformer-based models also established a new paradigm in the task of automatic speech recognition by introducing Wav2Vec 2.0\footnote{For brevity, we will omit the version number in the following text and denote this model only as \texttt{Wav2Vec}.} model \cite{baevski2020wav2vec}. 
It is a Transformer encoder pre-trained to reconstruct the corrupted signals. 
The raw audio signal input is processed by the model into a sequence of frame-level \textit{contextualized speech representations} encoding individual audio frames within their context. 
The training of Wav2Vec ASR models typically consists of two main phases: self-supervised pre-training and supervised fine-tuning. When fine-tuning, the model is supplemented with the final Connectionist Temporal Classification (CTC) layer \cite{Graves:06icml}, in which the most probable sequence of text tokens (i.e., the predicted transcription) is decoded.

\section{Selected languages}
To simplify the experiments, we chose 3 languages well-represented in both fine-tuning datasets we were working with (see Sec. \ref{sec:ft}): English, German, and Czech. They all belong to the Indo-European language family, Czech belongs to the Balto-Slavic branch while English and German are Germanic languages with higher mutual lexical similarity. We chose these languages for several reasons: (1) German is a language whose phrases intertwine throughout the MALACH archive, so it should be included; (2) English is the most represented and studied part of both datasets; and (3) Czech is well-represented in both datasets while it is an example of language from a completely different language branch. This language selection allows us to test various language combinations and see whether lexical similarity will be somehow reflected in results from multilingual models. We didn't include more languages to keep the number of possible language combinations reasonably small for experimenting and the results interpretable.

\section{Pre-trained models}
We started from the Wav2Vec-base English model \cite{baevski2020wav2vec} pre-trained from 50k hours from Libri-light dataset \cite{kahn2020libri}. We used the model as a base for English monolingual ASR models. 
To ensure comparable results of our experiments, we adopted the exact same pre-training setup for all other models we pre-trained from scratch, and scaled pre-training data for the other two languages to the same amount. For Czech and German, we collected 50k hours of speech from public sources, mainly from the VoxPopuli dataset \cite{wang-etal-2021-voxpopuli} and a mix of self-crawled publicly available podcasts and audiobooks.

\begin{figure}[t]
  \centering
  \includegraphics[width=1\linewidth]{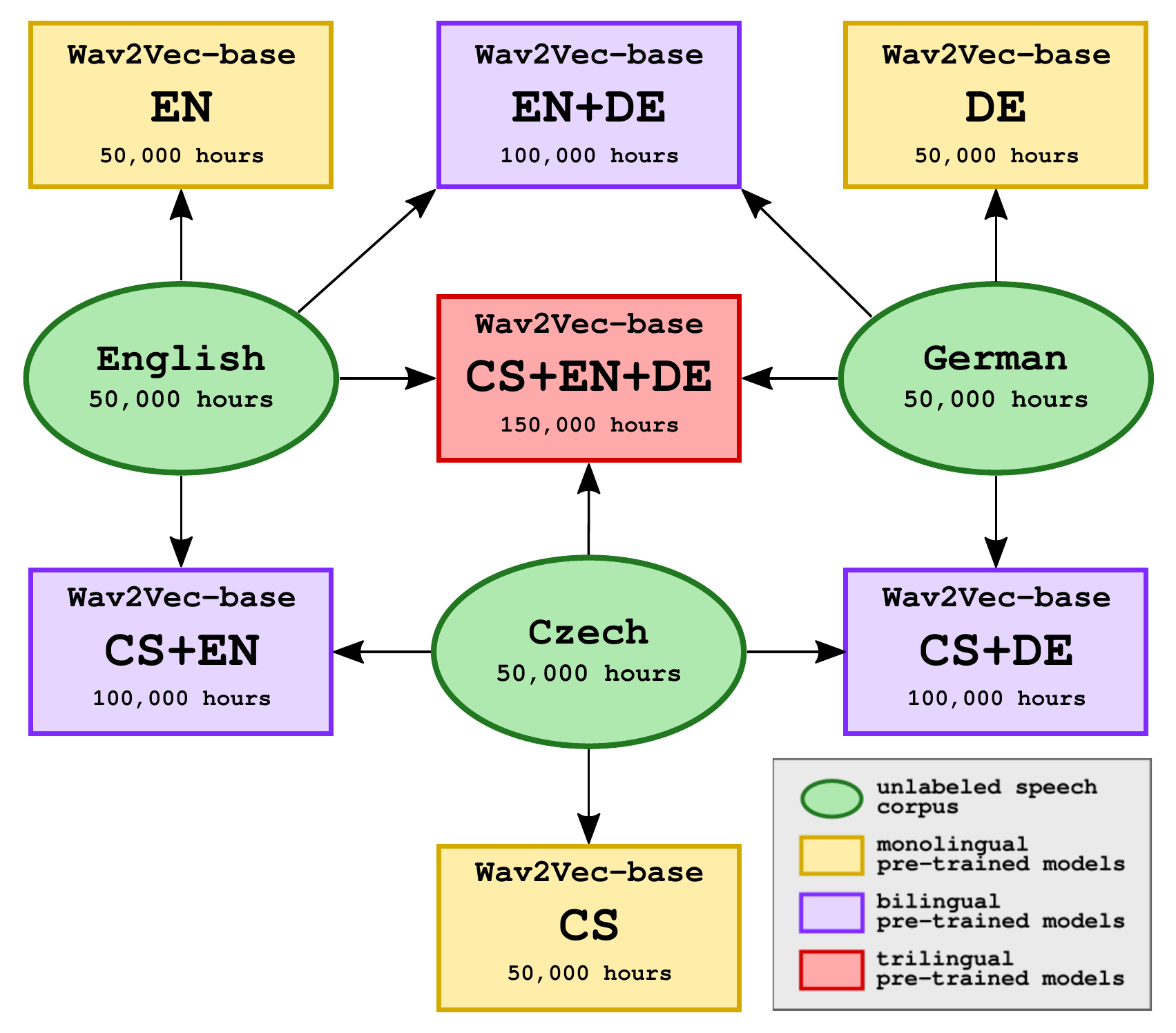}
  \caption{Scheme of pre-training mono-, bi- and trilingual Wav2Vec models.}
  \label{fig:scheme_pretrain}
\end{figure}

The scheme of pre-training our models is depicted in Fig. \ref{fig:scheme_pretrain}.
First, we pre-trained German and Czech monolingual Wav2Vec-base models from 50k hours of speech data each. Then, we prepared a mixture of all pairs of languages to train bilingual models from 100k hours each. Finally, we combine data from all three languages to pre-train a trilingual model from 150k hours of the equally balanced trilingual dataset. We experimented with three monolingual, three bilingual, and one trilingual model. Except for the monolingual English model, we pre-trained all models from scratch using the exact same setting as used for the English model, i.e., we trained the base model for 400 thousand steps with a batch size of about 1.6 hours. We used \texttt{Fairseq} tool\footnote{\url{https://github.com/pytorch/fairseq}} for pre-training all models. Pre-training of each model took approx. five days on a machine with eight NVIDIA A100 GPUs.
We are releasing all newly pre-trained models publicly to the research community\footnote{\url{https://huggingface.co/fav-kky}}.

\subsection{Large-scale multilingual models}
To compare our Wav2Vec models also with large-scale multilingual models, we selected \texttt{Wav2Vec-XLS-R-300M} \cite{babu22_interspeech} and \texttt{Whisper} \cite{radford2022whisper}. Wav2Vec-XLS-R-300M is a popular model pre-trained by Meta~AI on 128 languages and approximately 436 thousand hours of unlabeled speech data. We experimented with the 300M variant, which has more than $3\times$ more parameters than the Wav2Vec-base model.
Whisper is another popular model trained by OpenAI on 99 languages from 680,000 hours of multilingual and multitask labeled data. It is an encoder-decoder model already fine-tuned on multilingual ASR tasks by the authors, so it can also be used as a zero-shot speech recognizer without fine-tuning. We experimented with three sizes of the Whisper model: \texttt{base}, \texttt{small}, and \texttt{large}.

\section{Fine-tuning datasets}
\label{sec:ft}
We were experimenting with two multilingual datasets: CommonVoice and MALACH. 
For both datasets and all selected languages, we cleaned all transcripts by removing non-speech events and punctuation and mapping texts into lowercase. The data statistics are shown in Tab.~\ref{tab:ASRstats}.

\setlength{\tabcolsep}{0.2em}
{\renewcommand{\arraystretch}{1.1}%
\begin{table}[htb]
  \caption{Fine-tuning datasets. We show the number of hours, words in transcripts (in thousands), and the average length of train/dev/test segments of audio (in seconds). Note that these statistics are from balanced trilingual datasets we used in this paper; the full datasets contain more data.}
  \label{tab:ASRstats}
  \centering
  \begin{tabular}{l|lrrrcrrrcrrr}
    \toprule
     & & \multicolumn{3}{c}{English} & & \multicolumn{3}{c}{German} & & \multicolumn{3}{c}{Czech} \\
    \cline{3-5} \cline{7-9} \cline{11-13} 
     & & train & dev & test & & train & dev & test & & train & dev & test \\
     \midrule
    
    \parbox[t]{2mm}{\multirow{3}{*}{\rotatebox[origin=c]{90}{\tiny{CommonVoice}}}} & \textit{\# hours} & 25.0 & 27.2 & 27.0 & & 25.0 & 27.4 & 27.4 & &  25.0 & 11.6 & 11.5 \\
    & \textit{\# words} & 161 & 157 & 153 &  & 150 & 150 & 148 &  & 149 & 67 & 65 \\
    & \textit{avg-len} & 5.7 & 6.0 & 5.9 &  & 5.7 & 6.1 & 6.1 &  & 4.8 & 4.6 & 4.6 \\
    
    \midrule
    
    \parbox[t]{2mm}{\multirow{3}{*}{\rotatebox[origin=c]{90}{\tiny{MALACH}}}} & \textit{\# hours} &  85.0 & 9.2 & 4.3 &  & 85.0 & 33.0 & 80.8 &  & 85.0 & 19.2 & 9.0 \\
    & \textit{\# words} & 669 & 73 & 36 &  & 632 & 251 & 563 &  & 600 & 137 & 63 \\
    & \textit{avg-len} & 26.8 & 24.0 & 5.3 &  & 20.3 & 20.3 & 1692 &  & 22.9 & 22.9 & 10.7  \\
   
    \bottomrule
  \end{tabular}
\end{table}
}

\subsection{CommonVoice}
The CommonVoice is a crowdsourced dataset collected by Mozilla \cite{commonvoice:2020}. We used corpus version 16.0 containing 19,673 validated hours in 120 languages. English portion contains 1,718 hours, German 912, and Czech 26 hours of training data. Since a mixture of these datasets is highly unbalanced and every mixed-language fine-tuning we ran with the full datasets ended in favor of English ASR with poor performance for the other two languages, we decided to balance the dataset equally and use only a randomly selected subset with 25 hours of training data per language.
With this significant reduction of English and German training data, we ensure equal importance and fair conditions for all languages during the training.
The development and test splits were used completely without any change.

\subsection{MALACH}
The full MALACH archive is a monumental collection with over 100,000 hours of interviews in 32 languages. About half of the archive is in English, although most English interviews are given by non-native speakers. The annotated English part contains 375 hours of speech, the German part almost 2,000 hours, and the Czech part 130 hours. Similarly to the CommonVoice dataset, we decided to balance this dataset and use an equal amount of training data for each language. The smallest training split is in the Czech part with about 87 hours of training data, so we randomly selected a subset with 85 hours of training data per language. We used full development and test splits without any additional changes. The test part had no speaker overlaps with train or development parts in all languages.

For English and Czech, we used datasets released under the Linguistic Data Consortium (LDC) -- English \cite{MALACHen} and Czech \cite{MALACHcz}. We adopted the same train-dev-test splits as in \cite{10.1007/978-3-030-87802-3_50} and segmented train and development parts using time labels from the annotations into segments not exceeding 30 s, which is a reasonable limit of input examples during training due to GPU memory limits. The test parts for these two languages were already cleaned and contained only selected shorter segments (usually covering the maximum length of a single speaker's utterance without overlaps).
As we found in \cite{10.1007/978-3-031-16270-1_25}, the Czech MALACH transcripts contain a mix of formal and colloquial Czech, causing a mismatch between train and test data, so we converted all Czech training transcripts into formal Czech to close the gap.

For German, we adopted the same data splitting and pre-processing as in \cite{lehecka23_interspeech} with additional removing of the non-speech token \textit{ah} from transcripts as we observed inconsistent annotations of this token. We used the full unsegmented test split without any further segmentation of filtration, so the recordings in the German test dataset are much longer and less clean than test data from other languages.

\section{Experiments}
We fine-tuned all Wav2Vec models with the same setting as the base model in \cite{baevski2020wav2vec}, i.e., we trained for 80 thousand steps with a batch size of about 26 minutes per step, and the learning rate warmed up over the first 8\,000~steps to a maximum value of $2\times10^{-5}$, where it was held for the next 32\,000~steps, and finally decayed exponentially to zero. The weights of the feature encoder were frozen for the first 10\,000~steps of the fine-tuning.

Whisper models were fine-tuned differently because we observed some overfitting tendencies. For each model size, dataset, and language, we run 4 different fine-tunings with learning rates set to $1\times10^{-5}$, $5\times10^{-5}$, $1\times10^{-4}$ and $5\times10^{-4}$. We trained all models for 10 epochs with a batch size of 32 and measured the error rate after each epoch on the development dataset. Finally, we chose a checkpoint with the lowest error rate for evaluation.
We fine-tuned Wav2Vec models with \texttt{Fairseq} and Whisper with the \texttt{Transformers} library\footnote{\url{https://github.com/huggingface/transformers}}, both on a machine with eight NVIDIA A100 GPUs. The training took approx. 12 hours (Wav2Vec-base), 30 hours (Wav2Vec-XLS-R), resp. less than 3 hours (Whisper).

We compared models in terms of word error rate (WER). Since all transcripts were in lowercase and cleaned from punctuation, our fine-tuned models cannot predict punctuation or upper-cased characters, so we did not consider casing and punctuation differences with the reference as errors.

\setlength{\tabcolsep}{0.6em}
{\renewcommand{\arraystretch}{1.0}%
\begin{table*}[t]
  \caption{WER $[\%]$ for monolingual and various multilingual ASR models evaluated on relevant test splits of two multilingual datasets: CommonVoice and MALACH. We use ISO 639-1 codes to denote individual languages: CS=Czech, EN=English, and DE=German.}
  \label{tab:results}
  \centering
  \begin{tabular}{clccllccccccc}
    \toprule
    & & \# of & \# of & Pre-training & Fine-tuning & \multicolumn{3}{c}{CommonVoice} & & \multicolumn{3}{c}{MALACH} \\
    \cline{7-9} \cline{11-13} 
    Row & Model type & params & langs & langs & langs & CS & EN & DE & & CS & EN & DE \\
    \midrule
    1 & Wav2Vec-base & 95M & 1 & CS/EN/DE & CS/EN/DE & \textbf{11.36} & \textbf{34.07} & \textbf{17.23} & & \textbf{11.19} & \textbf{19.89} & \textbf{20.55} \\
    \midrule
    2 & Wav2Vec-base & 95M & 2 & CS+EN     & CS/EN    & 15.08 & 37.24 &     - & & 13.62 & 21.93 &     - \\
    3 & (bilingual) & & &                     & CS+EN    & 14.99 & 36.66 &     - & & 14.89 & 23.12 &     - \\
    \cline{5-13}
    4 & & & & CS+DE               & CS/DE    & 14.97 &     - & 21.16 & & 13.16 &     - & 24.49 \\
    5 & & & &                     & CS+DE    & 14.69 &     - & 21.19 & & 14.20 &     - & 25.10 \\
    \cline{5-13}
    6 & & & & EN+DE               & EN/DE    &     - & 36.71 & 20.16 & &     - & 21.07 & 23.39 \\
    7 & & & &                     & EN+DE    &     - & 36.14 & 20.73 & &     - & 22.53 & 24.70 \\
    \midrule
    8 & Wav2Vec-base & 95M & 3 & CS+EN+DE & CS/EN/DE & 16.52 & 38.70 & 21.72 & & 14.19 & 21.66 & 24.87 \\
    9 & (trilingual) & & &                     & CS+EN    & 16.80 & 38.16 &     - & & 15.66 & 22.35 &     - \\
    10 & & & &                    & CS+DE    & 16.47 &     - & 22.11 & & 15.49 &     - & 26.56 \\
    11 & & & &                    & EN+DE    &     - & 38.09 & 22.30 & &     - & 22.74 & 26.41 \\
    12 & & & &                    & CS+EN+DE & 16.97 & 38.35 & 22.86 & & 16.82 & 24.29 & 28.49 \\
    \midrule
    \midrule
    13 & Wav2Vec-XLS-R & 315M & 128 & see \cite{babu22_interspeech} & CS/EN/DE & \textbf{11.37} & 26.96 & 14.27 & & 12.46 & 17.23 & 20.11 \\
    14 & Whisper-base & 74M & 99 & see \cite{radford2022whisper} & - & 70.62 & 25.35 & 28.66 & & 58.39 & 25.01 & 34.03 \\
    15 &  &  &  &  & CS/EN/DE & 26.69 & 24.81 & 27.42 & & 18.38 & 40.44  & 30.88 \\
    16 & Whisper-small & 244M & 99 & see \cite{radford2022whisper} & - & 38.03 & 17.07 & 15.10 & & 35.69 & 20.11 & 26.11 \\
    17 &  &  &  &  & CS/EN/DE & 18.51 & 29.27 & 15.74 & & 12.99 & 64.08  & 36.04 \\
    18 & Whisper-large & 1,550M & 99 & see \cite{radford2022whisper} & - & 14.23 & \textbf{11.67} & \textbf{7.29} & & 21.65 & \textbf{16.97} & 21.59 \\
    19 &  &  &  &  & CS/EN/DE & 15.14 & 13.59 & 8.95 & & \textbf{10.19} & 27.10  & \textbf{19.59} \\
    \bottomrule
  \end{tabular}
\end{table*}
}

Our results are tabulated in Tab.~\ref{tab:results}. In the first part of the table (rows 1-12), we are comparing monolingual Wav2Vec models with bilingual models in all possible language combinations, and with the trilingual model. We fine-tuned each pre-trained model on all combinations of languages that were possible. For \textit{monolingual fine-tuning} (i.e. we used data from one language only during the fine-tuning), we aggregated results into a single row and separated individual fine-tuning languages by a slash in the language columns. For example, on the 8th row, we took a trilingual pre-trained model (pre-training languages were CS+EN+DE), fine-tuned 6 different ASR models, one per each language (CS/EN/DE) and each dataset (CommonVoice, MALACH), and evaluated each model on corresponding test data. On the contrary, on the 12th row, we took the same pre-trained trilingual model and fine-tuned 2 different ASR models (one per dataset) from a mixture of 3 languages in the fine-tuning data (denoted as CS+EN+DE). In other words, we denote the joining of datasets for multilingual training by “+” (the resulting model is one multilingual model per dataset), and a set of language-specific monolingual training runs by “/”
resulting in one monolingual model per each language and dataset.

In the second part of Tab.~\ref{tab:results} (rows 13-19), we present results using large-scale multilingual models. It is important to consider individual models' sizes and fairly compare only models of similar sizes, so we also included the column with the number of trainable parameters in the table. We fine-tuned all 4 models (Wav2Vec-XLS-R and three sizes of Whisper) on all languages and datasets. The zero-shot performance of the Whisper models is reported on rows 14, 16, and 18.

\section{Discussion}

\begin{figure}[!b]
  \centering
  \includegraphics[width=\linewidth]{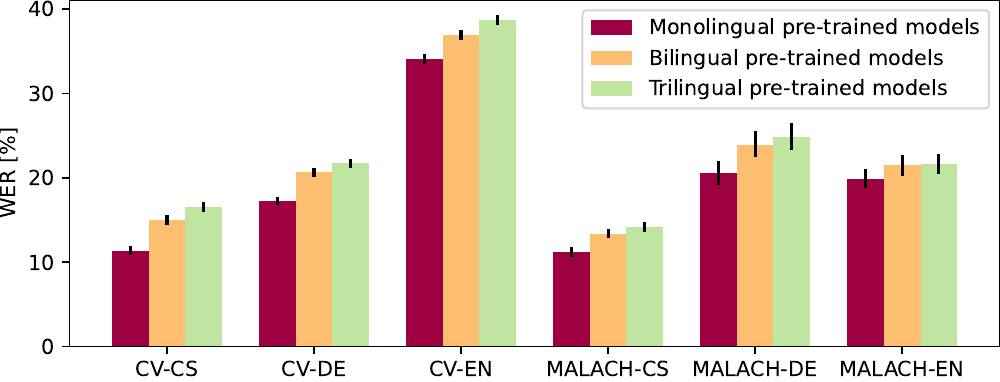}
  \caption{WER change after adding more languages into pre-training. We abbreviated CommonVoice to CV. We plot also confidence intervals at 95\% confidence level.}
  \label{fig:WER_langs_pt}
\end{figure}

Our results suggest that adding more languages into the pre-training phase while keeping the model at the same size did not bring any improvement for either dataset. On the contrary, we observed a trend in WER increasing when adding more languages. 
We plotted this interesting trend in Fig.~\ref{fig:WER_langs_pt}, where for each dataset and each language, we compared the monolingual model (row 1 in  Tab.~\ref{tab:results}), the average WER scored by bilingual models (rows 2, 4, and 6) and the trilingual model fine-tuned on a single language (row 8). This suggests that even when our dataset is multilingual and contains a lot of mixed-language sentences, the best we can do (when we want to keep the model reasonably small for production) is to train the monolingual model on each language separately from scratch.

It is worth noting that our best results in Tab.~\ref{tab:results} are not state-of-the-art results. We are just comparing different models under the same conditions. Significantly better results could be scored with monolingual models when using more training data and a language model in the CTC decoder \cite{picheny23_interspeech,lehecka23_interspeech}.

As for adding more languages into fine-tuning (assuming we already have a pre-trained multilingual model), we also did not observe any significant improvements. For the CommonVoice dataset, the results are moreover the same no matter whether we fine-tune a multilingual model or more monolingual models. For the MALACH dataset, we observed even an increase of WER after adding more languages into fine-tuning (compare e.g. rows 2 vs. 3, 4 vs. 5, 6 vs. 7, or 8-12 in Tab.~\ref{tab:results}).

Large-scale multilingual models could outperform monolingual Wav2Vec-base models, but only at the cost of a many times higher number of parameters, and thus higher computational complexity leading to higher price and carbon footprint for each decoded word. If we compare models with similar sizes as Wav2Vec-base (Whisper-base and -small), we can, again, observe the superiority of monolingual models. 
In several cases, the Whisper model tended to hallucinate after the fine-tuning, leading to higher WER than its zero-shot performance.
The Whisper models scored significantly better results on English and German CommonVoice datasets, which is probably due to the presence of the full datasets in the training process.

\section{Conclusions}

In this paper, we have presented a comparative analysis over a large set of experiments with different
language combinations in both pre-training and fine-tuning of
ASR systems based on Wav2Vec 2.0 models. We evaluated our models on two multilingual datasets and three languages. Our results suggest that monolingual Wav2vec models are, in most cases, superior to multilingual models. Only large-scale multilingual models, many times larger, can outperform monolingual Wav2Vec models, but only at the cost of much higher decoding complexity and carbon footprint for each transcribed word.

\section{Acknowledgements}
This research was supported by the Ministry of the Interior of the Czech Republic, project No. VJ01010108 and by the Ministry of Education, Youth and Sports of the Czech Republic through the e-INFRA CZ (ID:90254).

\bibliographystyle{IEEEtran}
\bibliography{mybib}

\end{document}